\documentclass[runningheads]{llncs}

\usepackage{eccv}
\usepackage{float}
\usepackage{makecell}
\usepackage{graphicx}
\usepackage{booktabs}

\usepackage{hyperref}
\usepackage{orcidlink}

\begin{document}

\title{LimeCross: Context-Conditioned Layered Image Editing with Structural Consistency} 
\titlerunning{LimeCross}

\author{
Ryugo Morita\inst{1} \and
Stanislav Frolov\inst{1} \and
Brian Bernhard Moser\inst{1} \and
Ko Watanabe\inst{1} \and
Riku Takahashi\inst{2} \and
Issey Sukeda\inst{3} \and
Andreas Dengel\inst{1}
}

\authorrunning{R.Morita et al.}

\institute{
RPTU Kaiserslautern-Landau \& DFKI GmbH, Kaiserslautern, Germany \and
Faculty of Science and Engineering, Hosei University, Tokyo, Japan \and
EQUES, Tokyo, Japan}
\maketitle

\begin{figure}[H]
    \centering
    \includegraphics[width=\linewidth]{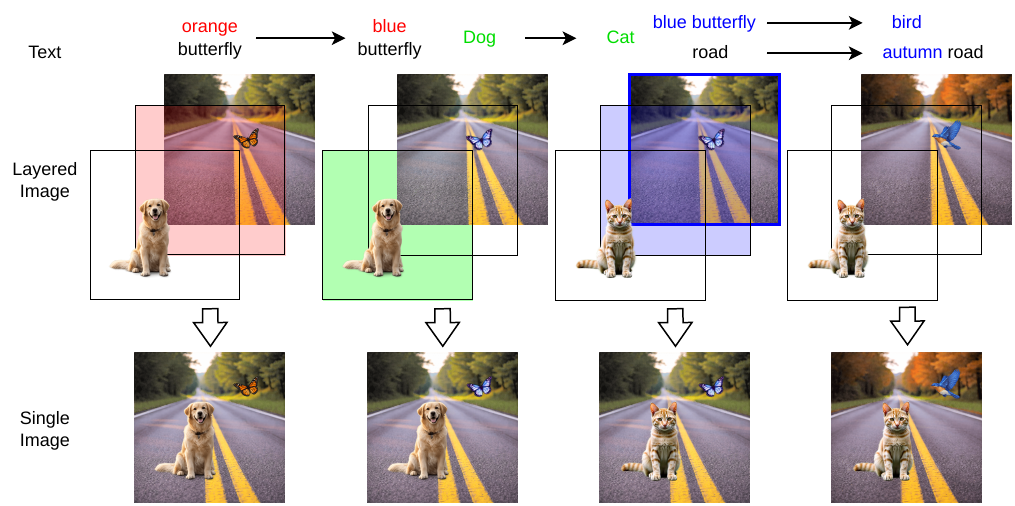}
    \caption{
    LimeCross enables text-guided image editing within a layered RGBA image set, modifying selected layers while preserving all others as reusable assets.
    By leveraging contextual cues from the remaining layers, it maintains cross-layer consistency in illumination, contact, and appearance, making it well-suited for layered creation workflows that require independent yet coherent control of multiple image elements.
    }
  \label{fig:teaser}

\end{figure}

\begin{abstract}
Layered image assets are widely used in real-world creative workflows, enabling non-destructive iteration and flexible re-composition. Recent advances in layered image generation and decomposition synthesize or recover layered representations, yet controllable editing of layered images remains challenging.
Manual editing requires careful coordination across layers to maintain consistent illumination and contact, while AI-based pipelines collapse layers into a flattened image for editing, then decompose them again, introducing background-to-foreground leakage and unstable transparency.
To address these limitations, we propose LimeCross, a training-free context-conditioned layered image editing framework that edits user-selected RGBA layers according to text while keeping the remaining layers unchanged.
It leverages contextual cues from other layers using a bi-stream attention mechanism to preserve cross-layer consistency, while explicitly maintaining layer integrity to prevent the contamination of edited layers.
To evaluate our approach, we introduce LayerEditBench, a benchmark of 1500 layered scenes with paired source/target prompts, along with evaluation protocols that assess both edit fidelity and alpha channel stability. 
Extensive experiments demonstrate that LimeCross improves layer purity and composite realism over strong editing baselines, establishing context-conditioned layered editing as a principled framework for controllable generative creation.
\end{abstract}

\section{Introduction}
\label{sec:introduction}
Layered image assets are crucial to real-world creative workflows in graphic design, animation, and advertising, where foreground objects and background plates are manipulated as independent layers.
This representation enables non-destructive, iterative editing, preserving assets as reusable, recomposable components rather than committing to a single flattened composite.
By explicitly separating scene elements, layered representations encode structural information such as depth, occlusion, and object interactions, enabling structured scene manipulation.
Motivated by these workflows, recent advances in text-to-layered image generation~\cite{huang2025dreamlayer, liu2025omnipsd, pu2025art} and layered image decomposition~\cite{liu2025controllable, yin2025qwen, suzuki2025layerd} have enabled the synthesis and recovery of layered RGBA representations from text and RGB inputs. Despite recent advances in layered generation and decomposition, representation alone is not the end goal.
The central challenge is to transform layered representations into actionable control: enabling context-aware editing of individual layers while preserving cross-layer consistency and structural integrity.
In this sense, layering is valuable not merely as a representation but as a structural foundation for coherent and controllable scene manipulation.

In practice, layered images are edited either manually or via a flatten-then-edit pipeline, where layers are first composited into a single RGB image, edited using a text-guided method~\cite{kulikov2025flowedit,zhu2025kv,xie2025dnaedit}, and then decomposed again.
Manual editing preserves structure but makes it difficult to consistently model illumination, contact, and inter-layer interactions.
Flatten-then-edit pipelines leverage context but fundamentally destroy layer independence, requiring re-decomposition after each modification and introducing cross-layer leakage, unstable transparency, and brittle behavior under occlusion.
Several recent works have explored editing beyond flattened RGB images, either by operating directly in RGBA space~\cite{dai2025trans} or by accepting multi-layer inputs~\cite{liu2025magicquillv2}.
Trans-Adapter~\cite{dai2025trans} enables inpainting-based editing of a single RGBA image, but treats it in isolation and cannot incorporate interactions with other layers.
MagicQuillV2~\cite{liu2025magicquillv2} supports multi-layer inputs during editing, yet it produces a flattened output, which requires re-decomposition for subsequent edits and limits compatibility with iterative workflows.
As a result, existing methods do not maintain layered representations throughout editing and fall short of providing a practical solution for controllable layered editing that preserves both layer integrity and cross-layer consistency.

To address this gap, we propose LimeCross, a training-free context-conditioned layered image editing framework.
Given multi-layer RGBA inputs, a user selects a target layer and edits it with text while keeping all other layers unchanged.
The central challenge is to exploit contextual information from the remaining layers without contaminating the edited layer itself.
Our key insight is to \emph{read context without copying it}: the editable layer attends to other layers to infer scene-level cues, while preventing contextual appearance from being written back into the RGBA output.
Concretely, we introduce a bi-stream attention mechanism that controls contextual interactions at inference time, preserving cross-layer consistency while maintaining layer independence, all without additional training.

To enable systematic evaluation in layered settings, we introduce LayerEditBench, a benchmark of 1500 layered scenes with paired source and target prompts, together with evaluation protocols that measure both edit fidelity and layer integrity.
Extensive experiments demonstrate that LimeCross improves layer purity and composite realism over strong editing baselines in both single-layer and iterative multi-layer editing settings. These findings suggest that treating layered editing as a structural representation rather than a post-hoc decomposition provides a principled foundation for more controllable and semantically grounded generative creation.
Our contributions are summarized as follows:

\begin{itemize}
    \item We formulate context-conditioned layered image editing for layered RGBA inputs, where a user-selected target layer is edited via text while all other layers remain unchanged, and we define layer-integrity criteria that capture leakage and alpha channel stability.

    \item We propose LimeCross, a training-free framework that performs context-conditioned layered editing via a bi-stream attention mechanism, enabling the selected RGBA layer to attend to remaining layers while preserving strict layer separation and preventing write-back leakage.
    
    \item  We introduce LayerEditBench with 1500 layered scenes and paired source--target prompts, together with evaluation protocols for both single-layer and iterative multi-layer editing that measure edit fidelity and alpha channel stability, and we demonstrate consistent improvements over baselines.
\end{itemize}

\section{Related work}
\label{sec:related}

\subsection{Layered Image Generation and Decomposition}
Prior work on layered images can be categorized into
(i) layered image generation, which synthesizes RGBA layer sets from text,
and (ii) layered decomposition, which recovers layered representations from RGB images or videos.

\noindent
\textbf{Layered Image Generation} aims to synthesize multi-layer RGBA representations from text, where scene elements are explicitly separated into RGBA layers.
Existing approaches rely on transparency-aware designs, including special masking~\cite{zhang2023text2layer, huang2024layerdiff, fontanella2024generating, kang2025layeringdiff} or alpha-aware autoencoders~\cite{zhang2024transparent, dalva2024layerfusion, huang2025psdiffusion, pu2025art}. Training-free alternatives~\cite{morita2025tkg, quattrini2024alfie, zou2025zero} eliminate additional data collection or model fine-tuning.
More recent work extends layered generation from single assets to multi-layer RGBA outputs.
Some methods explicitly generate coupled foreground--background image~\cite{zhang2023text2layer,kang2025layeringdiff,nagai2025taue,huang2025psdiffusion,dalva2024layerfusion} and video~\cite{dong2025video,ji2025layerflow,wang2025transpixeler,chen2025transanimate,cen2025layert2v,bai2025layer} layers,
while others aim to synthesize a larger set of layers in a single forward pass, exposing multiple editable components for downstream manipulation~\cite{morita2023interactive,fontanella2024generating,chen2025prismlayers,song2025layertracer,pu2025art,huang2025dreamlayer,liu2025omnipsd,morita2026lgtm}.

\noindent
\textbf{Layered Image Decomposition} aims to recover layered assets from a composite image or video.
These approaches typically predict per-layer colors and alpha mattes under an alpha-compositing constraint, incorporating generative priors, inpainting objectives~\cite{chen2025inpainting,tudosiu2024mulan,suzuki2025layerd,niedecomposition}, or diffusion-based refinement~\cite{ao2025open,yang2025generative,liu2025controllable,yin2025qwen} to handle occlusions and missing content.
Recent work further extends layered decomposition to video by addressing temporal consistency across frames~\cite{lee2025generative}.

Despite this progress, existing methods treat layered representations as an end product.
In practical creative workflows, however, layering serves as a structural basis for subsequent manipulation.
This motivates our downstream task of context-aware layer editing, which requires modifying a selected layer while preserving cross-layer consistency and layer integrity, an aspect that remains underexplored in existing text-guided editors.

\subsection{Text-guided Image Editing}
Text-guided image editing modifies an input image according to a user instruction while preserving irrelevant content.
While training-based approaches achieve strong performance with paired edit data~\cite{labs2025flux,brooks2023instructpix2pix}, many recent works focus on training-free methods that reuse pretrained generative models at inference time. The benefit of training-free methods is their practicality and ability to adapt to new instructions without additional fine-tuning.

Training-free editors can broadly be categorized into inversion-based and inversion-free approaches.
Inversion-based methods first map the source image into the latent or noise space of a pretrained model and then reconstruct an edited result under a target prompt, improving locality and semantic control via attention~\cite{hertz2022prompt,mokady2023null,rout2024semantic,wang2024taming,yan2025eedit} or guidance manipulation~\cite{ju2023direct,ma2025adams,xu2025unveil,xie2025dnaedit,ronai2025flowopt}.
In contrast, inversion-free methods directly modify the sampling trajectory without explicit inversion. These include consistency-based sampling~\cite{xu2023inversion} and velocity-based formulations such as FlowEdit~\cite{kulikov2025flowedit}, with subsequent refinements for stability and content preservation~\cite{kim2025flowalign,yang2025fia,zhu2025kv,ouyang2025lore}.

Motivated by practical layered workflows, recent works have begun to move beyond flattened RGB editing toward RGBA-aware editing.
Trans-Adapter~\cite{dai2025trans} edits a single RGBA object but treats it in isolation.
MagicQuillV2~\cite{liu2025magicquillv2} accepts multi-layer inputs yet operates on a composited representation and produces a flattened output, requiring re-decomposition for layered reuse.
Consequently, existing methods either collapse layered inputs into a flattened representation or fail to preserve explicit layer separation during editing. 

In contrast, we aim to maintain the original layered structure throughout the editing process, enabling context-aware modification of a selected layer while preserving cross-layer information and enabling stable iterative multi-layer editing.

\begin{figure}[t]
  \centering
  \includegraphics[width=\linewidth]{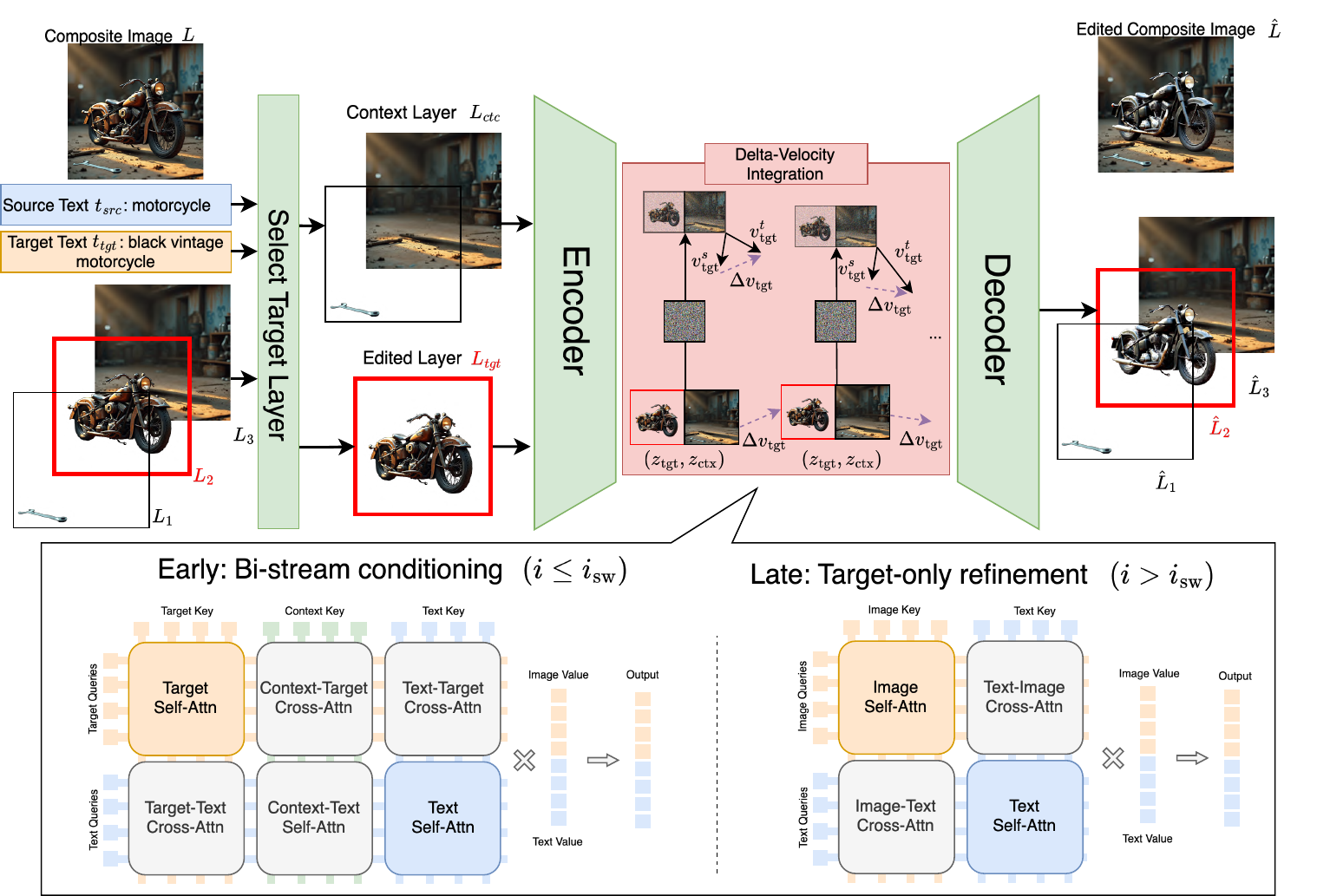}
\caption{
Given layered RGBA inputs, we select a target layer (red box) and construct an \emph{opaque context} by compositing all non-target layers.
Both are encoded into latents and packed into two token streams $(z_{\mathrm{tgt}}, z_{\mathrm{ctx}})$.
Editing is performed via delta-velocity integration: we evaluate source/target velocities and update only the target stream using
$\Delta v_{\mathrm{tgt}} = v^{t}_{\mathrm{tgt}} - v^{s}_{\mathrm{tgt}}$,
while context tokens are never updated or decoded.
In early steps, bi-stream attention establishes cross-layer cues.
After $i_{\mathrm{sw}}=\lfloor \rho T \rfloor$, we switch to target-only refinement,
disabling context interactions to stabilize fine details and alpha boundaries.
}
  \label{fig:model}
\end{figure}

\section{Methods}
\label{sec:methods}
As shown in Fig.~\ref{fig:model}, we consider a layered scene represented by $K$ RGBA layers
$\mathcal{L}=\{L_1,\dots,L_K\}$, where $L_k\in[-1,1]^{4\times H\times W}$,
and the stacking order (front-to-back) of the layers is assumed to be known.
Given a target layer index $\ell$, which specifies the user-selected editable layer $L^{(\ell)}$, a source prompt $p_{\text{src}}$, and a target prompt $p_{\text{tgt}}$,
our goal is to output an edited RGBA layer $\hat{L}^{(\ell)}$ such that
(i) $\hat{L}^{(\ell)}$ follows the instruction $p_{\text{src}}\rightarrow p_{\text{tgt}}$,
(ii) all non-target layers remain unchanged, and
(iii) the edited layer is compatible with inter-layer interactions when re-composited, preserving contextual coherence (illumination/contact) without back-to-foreground leakage or unstable transparency.
We perform editing with a two-stage schedule: we use bi-stream context conditioning in early steps and switch to target-only refinement in later steps.
The switching step is defined as $i_{\mathrm{sw}}=\lfloor \rho T \rfloor$, where $T$ is the total number of steps and $\rho\in(0,1)$ is the switching ratio.

\subsection{Latent Representation of Foreground and Context}
\label{subsec:latent_fg_ctx}
We perform editing in the latent space of an RGBA-capable VAE~\cite{wang2025alphavae},
with encoder $\mathcal{E}$ and decoder $\mathcal{D}$, conditioning the target layer on a context latent from the \emph{non-target} layers.
We encode the target layer into a target latent
\begin{equation}
x_{\mathrm{tgt}} = \mathcal{E}\!\left(L^{(\ell)}\right) \in \mathbb{R}^{C\times h\times w},
\end{equation}
where $h=H/s$ and $w=W/s$.
We denote VAE latents on the spatial grid by $x\in\mathbb{R}^{C\times h\times w}$, and use $\mathrm{pack}(\cdot)$ to reshape them into token sequences $z=\mathrm{pack}(x)\in\mathbb{R}^{N\times d}$ for transformer processing with $\mathrm{unpack}(\cdot)$ as the inverse.

To condition the edit on the remaining layers without copying their appearance into the edited RGBA output, we construct an \emph{opaque} context scene by compositing all layers \emph{except} the edited one.
Let $\mathcal{L}_{\neg \ell}$ denote the ordered set of non-target layers in the known stacking order (front-to-back),
and let $\mathbf{c}_b$ be an \emph{opaque} background plate (e.g., the RGB of the backmost layer).
For each layer $L_{k}=(\mathbf{c}_k,\alpha_k)$, we accumulate RGB contributions using standard alpha-over in RGB while tracking cumulative coverage $\tilde{\alpha}$:
\begin{align}
\tilde{\mathbf{c}}_{0} &= \mathbf{0}, \quad \tilde{\alpha}_{0} = 0,\\
\tilde{\mathbf{c}}_{t+1} &= \tilde{\mathbf{c}}_{t} + \mathbf{c}_{i_t}\odot \alpha_{i_t}\odot(1-\tilde{\alpha}_{t}),\\
\tilde{\alpha}_{t+1} &= \tilde{\alpha}_{t} + \alpha_{i_t}\odot(1-\tilde{\alpha}_{t}),
\end{align}
where $i_t$ indexes layers in $\mathcal{L}_{\neg \ell}$.
After processing all layers in $\mathcal{L}_{\neg \ell}$, we denote the final values by
$(\tilde{\mathbf{c}}_{\neg \ell}, \tilde{\alpha}_{\neg \ell})$.
We then obtain an opaque context image by filling uncovered regions with the background plate and fixing the output alpha to one:
\begin{equation}
\mathbf{c}_{\mathrm{ctx}} = \tilde{\mathbf{c}}_{\neg \ell} + \mathbf{c}_b \odot (1-\tilde{\alpha}_{\neg \ell}),
\qquad
S^{(\ell)} = (\mathbf{c}_{\mathrm{ctx}}, 1).
\end{equation}
Finally, we encode the context scene into a context latent
\begin{equation}
x_{\mathrm{ctx}} = \mathcal{E}\!\left(S^{(\ell)}\right)
\in \mathbb{R}^{C\times h\times w}.
\end{equation}

\subsection{Bi-stream Context Conditioning via Token Concatenation}
\label{subsec:bistream}
To expose contextual cues without flattening layers into a single RGB input, we provide the target layer and the context scene as two token streams and let the transformer perform cross-layer attention.
Following the MMDiT~\cite{esser2024scaling} implementation, we reshape each latent grid into a token sequence via $\mathrm{pack}(\cdot)$, with $\mathrm{unpack}(\cdot)$ as its inverse:
\begin{equation}
z_{\mathrm{tgt}}=\mathrm{pack}(x_{\mathrm{tgt}})\in \mathbb{R}^{N\times d},\quad
z_{\mathrm{ctx}}=\mathrm{pack}(x_{\mathrm{ctx}})\in \mathbb{R}^{N\times d}.
\end{equation}
We concatenate the streams along the token dimension
\begin{equation}
z = [z_{\mathrm{tgt}};\,z_{\mathrm{ctx}}] \in \mathbb{R}^{2N\times d},
\end{equation}
and feed $z$ to the pretrained transformer together with text conditioning.
In practice, we use separate image-id embeddings to distinguish the two streams.

Under standard self-attention, tokens interact via
\begin{equation}
\mathrm{Attn}(z)=\mathrm{softmax}\!\Big(\frac{QK^\top}{\sqrt{d}}\Big)V,\qquad
Q=zW_Q,\;K=zW_K,\;V=zW_V.
\end{equation}
Because $z=[z_{\mathrm{tgt}};z_{\mathrm{ctx}}]$, the attention weights form a block structure
\begin{equation}
\mathrm{softmax}\!\Big(\frac{QK^\top}{\sqrt{d}}\Big)=
\begin{bmatrix}
A_{\mathrm{tt}} & A_{\mathrm{tc}}\\
A_{\mathrm{ct}} & A_{\mathrm{cc}}
\end{bmatrix},
\end{equation}
where $A_{\mathrm{tc}}$ enables target tokens to attend to context tokens.
At each step, the re-noised context stream participates in the attention mechanism, while only the target stream is updated and decoded. 
Contextual tokens influence the editable stream solely through attention and are never directly updated, integrated, or decoded. 
This asymmetric design enables the model to infer scene-level cues such as illumination and contact without flattening the layered representation, while preserving strict architectural separation and preventing write-back leakage.

\subsection{LimeCross Trajectory by Delta-Velocity Integration}
We follow an inversion-free trajectory and update the editable target state using the difference between velocities conditioned on the source and target prompts.
For noise level $\tau \in [0,1]$, we define $\Phi_\tau(\cdot)$ as a linear forward noising operator in token space that mixes a clean stream with Gaussian noise $\epsilon \sim \mathcal{N}(0,I)$:
\begin{equation}
\Phi_\tau(z,\epsilon) = (1-\tau)\,z + \tau\,\epsilon.
\end{equation}
At a given step $\tau$, we form a noised source state for each stream via
\begin{equation}
z^{s}_{\mathrm{tgt}}(\tau)=\Phi_\tau(z_{\mathrm{tgt}},\epsilon_{\mathrm{tgt}}),\quad
z^{s}_{\mathrm{ctx}}(\tau)=\Phi_\tau(z_{\mathrm{ctx}},\epsilon_{\mathrm{ctx}}).
\end{equation}
We maintain an editable target state $z^{e}_{\mathrm{tgt}}(\tau)$.
As in FlowEdit~\cite{kulikov2025flowedit}, we define the \emph{target latent} for velocity evaluation by
\begin{equation}
z^{t}_{\mathrm{tgt}}(\tau)=z^{e}_{\mathrm{tgt}}(\tau)+z^{s}_{\mathrm{tgt}}(\tau)-z_{\mathrm{tgt}}.
\end{equation}
Here $z^{s}_{\mathrm{tgt}}(\tau)$ denotes the noised source stream, 
$z^{t}_{\mathrm{tgt}}(\tau)$ is the evaluation stream under the target prompt,
and $z^{e}_{\mathrm{tgt}}(\tau)$ is the editable state that is iteratively updated.
We use the same noise realization $\epsilon_{\mathrm{tgt}}$ for both source and target evaluations to isolate prompt-induced changes. We evaluate the target-stream velocities under the source and target prompts using the same re-noised context stream:
\begin{equation}
v^{s}_{\mathrm{tgt}} = f_\theta\!\left([z^{s}_{\mathrm{tgt}}(\tau);\; z^{s}_{\mathrm{ctx}}(\tau)], p_{\text{src}}, \tau\right),\quad
v^{t}_{\mathrm{tgt}} = f_\theta\!\left([z^{t}_{\mathrm{tgt}}(\tau);\; z^{s}_{\mathrm{ctx}}(\tau)], p_{\text{tgt}}, \tau\right).
\end{equation}
The editing direction is given by the delta-velocity on the target stream
\begin{equation}
\Delta v_{\mathrm{tgt}}(\tau)=v^{t}_{\mathrm{tgt}}-v^{s}_{\mathrm{tgt}}.
\end{equation}
We integrate this delta-velocity to update the editable target state
\begin{equation}
z^{e}_{\mathrm{tgt}}(\tau_{i+1}) = z^{e}_{\mathrm{tgt}}(\tau_i) + (\tau_{i+1}-\tau_i)\,\Delta v_{\mathrm{tgt}}(\tau_i),
\end{equation}
with a discrete schedule $\{\tau_i\}_{i=0}^{T}$ defined by the scheduler.

\subsection{Late-stage Target-only Refinement and Decoding}
\label{subsec:late_stage}
We adopt a two-stage editing schedule to balance contextual consistency and layer integrity.
In early steps, context conditioning helps establish coarse cross-layer alignment,
including illumination and global appearance.
In later steps, the model predominantly refines high-frequency details and transparency boundaries. Keeping cross-layer interactions at this stage over-impose context appearance onto the edited layer,
increasing appearance copying and destabilizing the alpha channel.

To mitigate this effect, LimeCross switches to a target-only refinement
after a predefined switching step $i_{\mathrm{sw}}$.
We set $i_{\mathrm{sw}}=\lfloor \rho T \rfloor$, where $\rho\in(0,1)$ controls the switching ratio.
For $i \ge i_{\mathrm{sw}}$, velocities are evaluated using only the target stream:
\begin{equation}
v^{s}_{\mathrm{tgt}} = f_\theta(z^{s}_{\mathrm{tgt}}(\tau_i), p_{\text{src}}, \tau_i), \quad
v^{t}_{\mathrm{tgt}} = f_\theta(z^{t}_{\mathrm{tgt}}(\tau_i), p_{\text{tgt}}, \tau_i),
\end{equation}
and the editable state $z^{e}_{\mathrm{tgt}}$ is updated using the same delta-velocity rule.

This stage-wise design preserves the global contextual alignment established in earlier steps,
while improving layer purity and alpha stability during refinement.
After the final step, we unpack the edited target tokens and decode
\begin{equation}
\hat{x}_{\mathrm{tgt}} = \mathrm{unpack}(z^{e}_{\mathrm{tgt}}), \quad
\hat{L}^{(\ell)} = \mathcal{D}(\hat{x}_{\mathrm{tgt}}).
\end{equation}
The resulting RGBA layer can be directly reused for compositing,
without requiring re-decomposition or per-image optimization.

\subsection{Sequential Multi-Layer Editing}
\label{subsec:multi}
We extend LimeCross to \emph{iterative layer-wise editing} over a user-selected set of layers
$\mathcal{T}\subseteq\{1,\dots,K\}$.
Unlike composited-output editors, LimeCross maintains an explicit layered representation throughout the editing process, enabling repeated refinements without re-flattening or re-decomposition.
For each editable layer $k\in\mathcal{T}$, we assume a source prompt $p_{\text{src}}^{(k)}$
and a target prompt $p_{\text{tgt}}^{(k)}$.

\subsubsection{Context construction excluding the edited layer}

To prevent copying appearance from context layers into the edited RGBA output, we construct the context scene by compositing all layers \emph{except} the currently edited one.
Let $\mathrm{Comp}(\cdot)$ denote the opaque compositing operator defined in Sec.~\ref{subsec:latent_fg_ctx}.
Given a current layered set $\tilde{\mathcal{L}}$ where already edited layers are replaced by their outputs, the context image for editing layer $k$ is
\begin{equation}
S^{(k)} = \mathrm{Comp}\!\left(\tilde{\mathcal{L}}_{\neg k}\right),
\end{equation}
where $\tilde{\mathcal{L}}_{\neg k}$ preserves the original stacking order.
The corresponding context latent is defined as
\begin{equation}
x_{\mathrm{ctx}}^{(k)} = \mathcal{E}(S^{(k)}).
\end{equation}
We then apply the same bi-stream token concatenation as in Sec.~\ref{sec:methods},
with $(x_{\mathrm{tgt}}^{(k)}, x_{\mathrm{ctx}}^{(k)})$.

\subsubsection{Back-to-front scheduling}

Layered scenes exhibit an asymmetric dependency: background layers establish global scene context, while foreground layers depend on them for consistent appearance and contact.
Motivated by this structure, we edit layers sequentially in a back-to-front order so that global context is stabilized before refining dependent foreground layers.
After each edit, the updated layer $\hat{L}^{(k)}$ replaces the original layer in $\tilde{\mathcal{L}}$, and subsequent edits operate on this updated layered set.
This iterative procedure reduces cross-layer ambiguity, limits error propagation across edits, and produces valid layered outputs at every step.
We initialize $\tilde{\mathcal{L}}\leftarrow \mathcal{L}$ and for each $k\in\mathcal{T}$ in back-to-front order,
we (i) build $S^{(k)}$ from $\tilde{\mathcal{L}}$, (ii) run the single-layer LimeCross routine
(Sec.~\ref{sec:methods}) to obtain $\hat{L}^{(k)}$, and (iii) update $\tilde{L}^{(k)}\leftarrow \hat{L}^{(k)}$.

\begin{figure}[t]
  \centering
  \includegraphics[width=\linewidth]{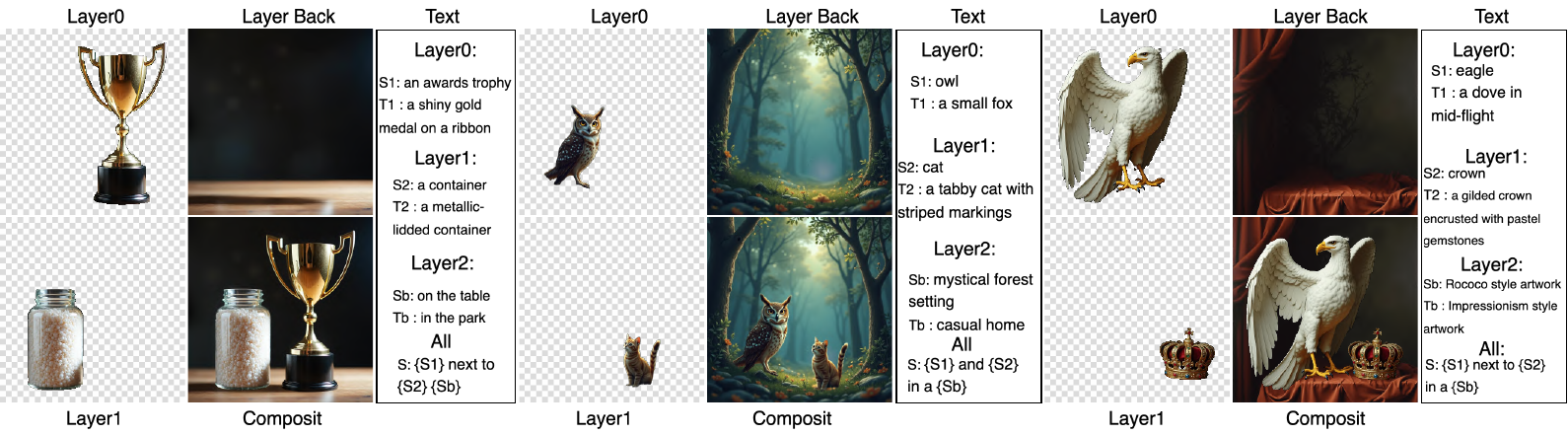}
  \caption{
    Sample layered assets from LayerEditBench, together with the corresponding source and target text for each layer.
    LayerEditBench includes a wide range of challenging layered editing scenarios, including object replacement, style changes, and atmosphere modifications.
    The benchmark features complex alpha channel structures, transparent objects, occlusions, and diverse artistic styles, enabling systematic evaluation of both edit fidelity and layer integrity.
    }
  \label{fig:bench}
\end{figure}

\section{Experiment}
\label{sec:experiment}
\subsection{LayerEditBench}
\label{subsec:layereditbench}
To systematically evaluate layered image editing beyond flattened RGB settings, we introduce LayerEditBench, a benchmark that assesses both edit fidelity and cross-layer integrity.
As illustrated in Fig.~\ref{fig:bench}, LayerEditBench contains 1500 layered scenes adapted from DreamLayer~\cite{huang2025dreamlayer} and PIE-Bench~\cite{ju2023direct}.
Each scene consists of a composite image and three ordered RGBA layers (two objects and background), together with layer-wise source/target prompt pairs specifying the intended edit for each layer.
The target prompts are created through a combination of human annotation and LLM~\cite{achiam2023gpt} assistance to ensure realism and diversity.
It covers a wide range of editing scenarios, including object replacement/addition and style/atmosphere changes.

For single-layer editing, we edit each layer independently, resulting in $1{,}500\times 3 = 4{,}500$ instances.
For iterative multi-layer editing, we edit all layer subsets of size 2 and 3 per scene using the sequential back-to-front procedure in Sec.~\ref{subsec:multi}, yielding $1{,}500\times\left(\binom{3}{2}+\binom{3}{3}\right)=6{,}000$ instances.

\subsection{Experimental Setup}
\label{subsec:setup}
We use FLUX.1-dev~\cite{flux2025} with an RGBA-capable VAE~\cite{wang2025alphavae} to obtain layered latent representations.
We use $T = 28$ steps and set the switching ratio to $\rho=0.5$.
Classifier-free guidance (CFG) is applied with a guidance scale of 1.5 for the source prompt and 5.5 for the target prompt.
All experiments are conducted on a single NVIDIA A100 GPU. For sequential multi-layer editing, we apply the sequential editing strategy described in Sec.~\ref{subsec:multi}.

For evaluation, we report HPSv2~\cite{wu2023human}, aesthetic score~\cite{schuhmann2022laion}, CLIP~\cite{hessel2021clipscore}, and ImageReward~\cite{xu2023imagereward} for edit quality, and PSNR, LPIPS, and MSE to measure preservation of non-edited regions.
Edited-region scores are computed within the alpha matte(s) of the edited layer(s), while preservation metrics are computed on pixels outside the edited region to assess how well unmodified content is maintained.
Overall scores are computed on the recomposited image.
For RGBA outputs with transparency, we composite results onto a constant mid-gray background (RGB $=0.5$ in $[0,1]$, i.e., $0$ in $[-1,1]$) before computing metrics.
To explicitly evaluate alpha-mask stability, we compute mFID-$\alpha$~\cite{morita2025tkg} between the output alpha masks and the corresponding input alpha masks.
For LimeCross, we directly use the decoded alpha channel of the edited RGBA layer.
For flattened-output baselines that do not produce layered outputs, we first apply Qwen-Image-Layered-Control~\cite{qwenlayeredcontrol2025} to decompose the edited composite image with text into layered representations and extract the predicted alpha mask of the edited layer.
mFID-$\alpha$ is then computed between these predicted alpha masks and the original input alpha masks to measure distributional consistency.

\subsection{Baselines and Adaptation to Layered Inputs}
\label{subsec:baselines}
We compare against KV-Edit~\cite{zhu2025kv}, a training-free text-guided editor designed for flattened RGB inputs, and MagicQuillV2~\cite{liu2025magicquillv2}, a training-based method that accepts multi-layer inputs but outputs a flattened image.
For baselines that require explicit spatial localization, we use the target layer alpha matte as the edit mask.
When editing a background layer or when objects are occluded, we remove pixels covered by layers in front (i.e., we subtract foreground mattes from the background mask) to avoid editing invisible regions.

For iterative multi-layer editing, we apply a sequential back-to-front strategy for the baselines, editing one layer at a time using its corresponding alpha matte as the mask.
After each edit, the updated result is used as the input context for subsequent edits, ensuring a consistent evaluation protocol across methods.
For MagicQuillV2, we feed the layer images in back-to-front stacking order and reorder the prompts accordingly.
For heavily occluded scenes, we apply a fixed tie-break rule that orders inputs by increasing visible-area ratio to reduce overlap conflicts.
Implementation details are provided in the supplementary material.

\begin{figure}[t]
  \centering
  \includegraphics[width=\linewidth]{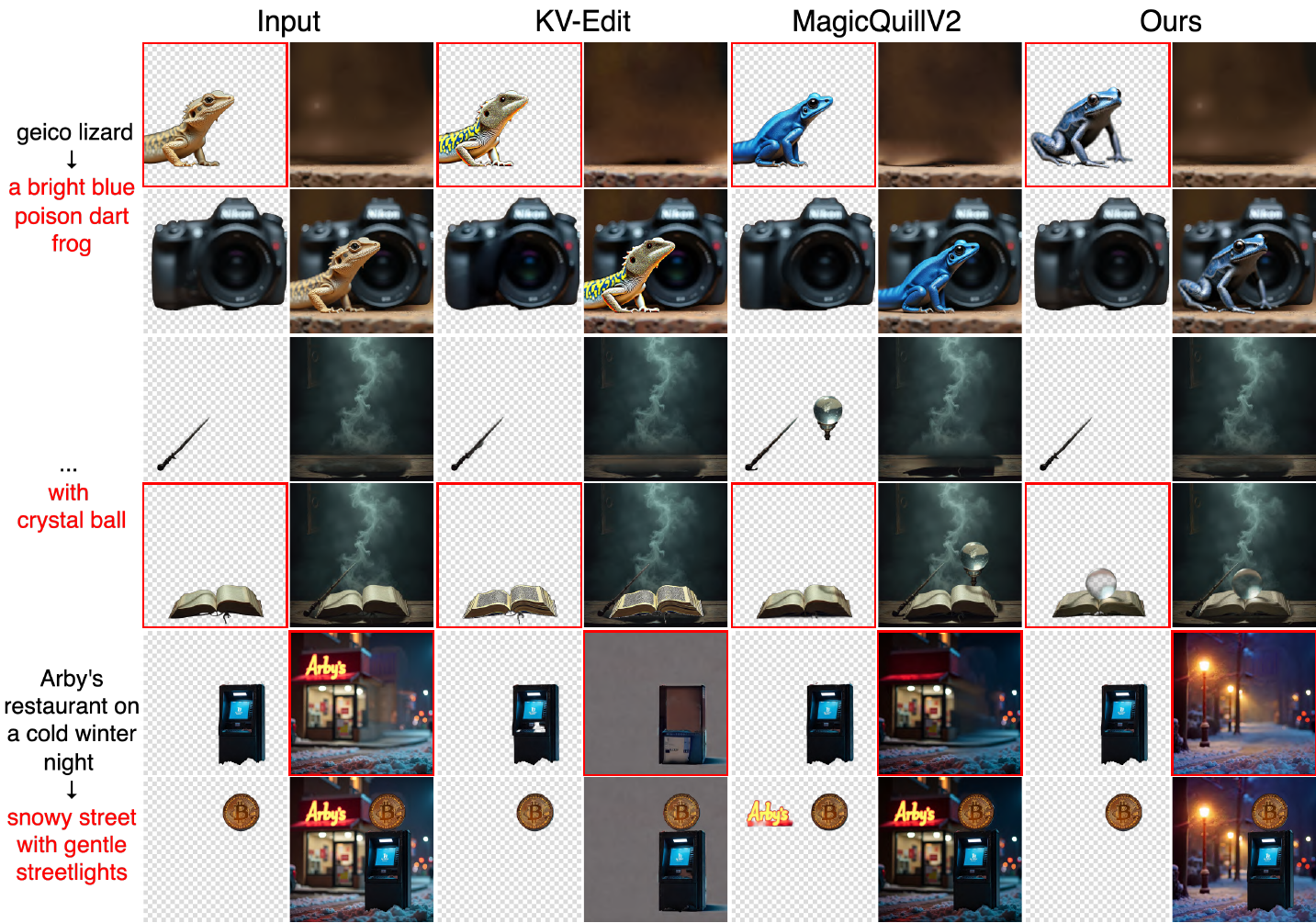}
  \caption{
      \textbf{Qualitative comparison for single-layer editing.}
    Flattened-output baselines produce visually plausible composites but fail to recover layered representations consistent with the original structure after decomposition, resulting in cross-layer leakage.
    In contrast, LimeCross edits the selected RGBA layer directly and preserves all non-target layers by construction, maintaining structural separation.}

  \label{fig:qualitative_single}
\end{figure}

\subsection{Qualitative Result}
\begin{figure}[t]
  \centering
  \includegraphics[width=\linewidth]{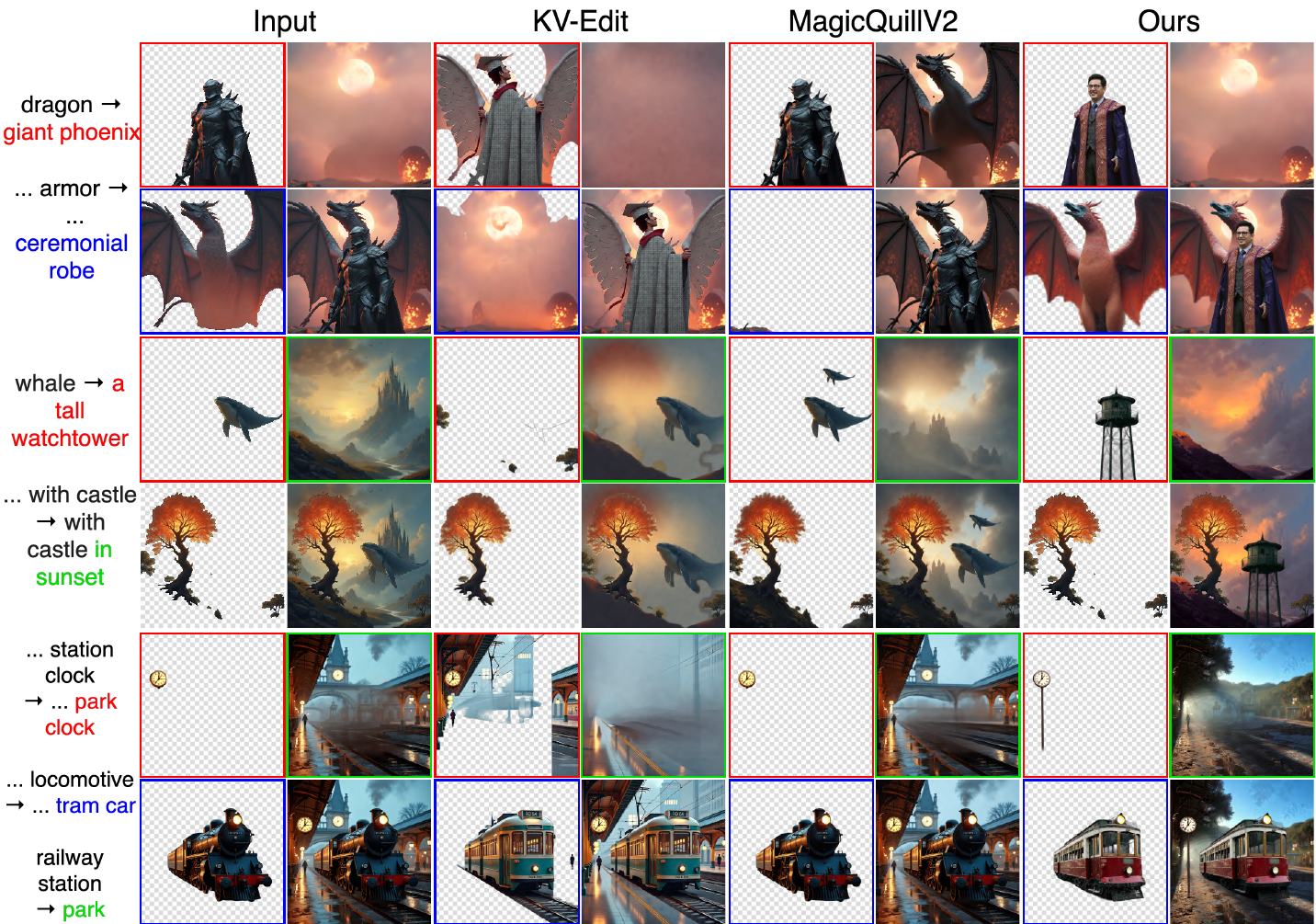}
    \caption{
    \textbf{Qualitative comparison for iterative multi-layer editing.}
    Under repeated edits, flattened-output methods accumulate errors, leading to structural drift and degraded scene consistency.
    Because each step requires re-flattening and reconstruction, artifacts compound across iterations.
    In contrast, LimeCross maintains explicit layered representations and updates only the designated layer at each step, enabling stable multi-step editing.
    }
  \label{fig:qualitative_multi}
\end{figure}

Fig.~\ref{fig:qualitative_single} and Fig.~\ref{fig:qualitative_multi} show qualitative results for our and existing methods in single-layer and iterative multi-layer editing.
For baselines that output flattened RGB images, we apply Qwen-Image-Layered~\cite{yin2025qwen} to visualize layered representations.

KV-Edit performs well when editing a clearly visible foreground object, but under overlap it struggles to infer occluded structures, leading to mixed or inconsistent reconstructions.
MagicQuillV2 achieves strong semantic alignment in single-image editing. However, its flattened-output design limits the benefit of layered inputs in iterative multi-layer settings. The originally present castle disappears after editing the whale, indicating structural drift and reduced layer confinement under heavy occlusion.(the second example of Fig.~\ref{fig:qualitative_multi})

In contrast, LimeCross performs editing directly on the selected RGBA layer while conditioning on contextual cues from other layers.
By keeping non-edited layers unchanged by construction, it preserves hidden structures and prevents cross-layer contamination.
As a result, LimeCross maintains structural consistency and stable separation even under occlusion and repeated multi-layer edits.

\begin{table*}[t]
\caption{
\textbf{Quantitative comparison on LayerEditBench.}
Edited-region metrics are computed within the alpha matte(s), and overall metrics on the recomposited image.
Preservation measures consistency outside the edited region.
For LimeCross, non-target layers remain unchanged by design, which yield trivial preservation scores (e.g., PSNR$=\infty$, LPIPS$=0$, MSE$=0$).
mFID-$\alpha$ denotes Fréchet Inception Distance computed on edited alpha masks to evaluate alpha stability.
}
\label{tab:all_settings}

\centering
\setlength{\tabcolsep}{1.5pt}
\renewcommand{\arraystretch}{1.25}
\scriptsize

\begin{tabular*}{\textwidth}{@{\extracolsep{\fill}}l|ccccc|cccc|ccc}
\toprule
\multicolumn{13}{c}{\textbf{Single-layer Image Editing}} \\
\midrule
Method
& \multicolumn{5}{c}{\textbf{Foreground}}
& \multicolumn{4}{c}{\textbf{Overall}}
& \multicolumn{3}{c}{\textbf{Preservation}} \\
\cmidrule(lr){2-6}\cmidrule(lr){7-10}\cmidrule(lr){11-13}
& \makecell{HPS\\{\tiny$\times 10^{2}$}$\uparrow$}
& AS$\uparrow$
& CLIP$\uparrow$
& IR$\uparrow$
& mFID-$\alpha$$\downarrow$
& \makecell{HPS\\{\tiny$\times 10^{2}$}$\uparrow$}
& AS$\uparrow$
& CLIP$\uparrow$
& IR$\uparrow$
& PSNR$\uparrow$
& \makecell{LPIPS\\{\tiny$\times 10$}$\downarrow$}
& \makecell{MSE\\{\tiny$\times 10^{2}$}$\downarrow$} \\
\midrule
KV-Edit
& 15.57 & 4.77 & 16.70 & -1.808 & 48.12
& 27.68 & 6.44 & 27.56 & 0.706
& 24.41 & 2.05 & 1.401 \\
MagicQuillV2
& 15.91 & 4.87 & 17.05 & -1.646 & 47.19
& 28.34 & 6.53 & 27.86 & 0.887
& 21.19 & 2.22 & 1.766 \\
\textbf{Ours}
& \textbf{21.19} & \textbf{4.89} & \textbf{23.52} & \textbf{-0.418} & \textbf{46.91}
& \textbf{28.75} & \textbf{6.55} & \textbf{28.69} & \textbf{1.116}
& $\infty$ & 0 & 0 \\
\midrule
\addlinespace[2pt]
\multicolumn{13}{c}{\textbf{Multi-layer Image Editing}} \\
\midrule
Method
& \multicolumn{5}{c}{\textbf{Edited Region}}
& \multicolumn{4}{c}{\textbf{Overall}}
& \multicolumn{3}{c}{\textbf{Preservation}} \\
\cmidrule(lr){2-6}\cmidrule(lr){7-10}\cmidrule(lr){11-13}
& \makecell{HPS\\{\tiny$\times 10^{2}$}$\uparrow$}
& AS$\uparrow$
& CLIP$\uparrow$
& IR$\uparrow$
& mFID-$\alpha$$\downarrow$
& \makecell{HPS\\{\tiny$\times 10^{2}$}$\uparrow$}
& AS$\uparrow$
& CLIP$\uparrow$
& IR$\uparrow$
& PSNR$\uparrow$
& \makecell{LPIPS\\{\tiny$\times 10$}$\downarrow$}
& \makecell{MSE\\{\tiny$\times 10^{2}$}$\downarrow$} \\
\midrule
KV-Edit
& 21.75 & 6.33 & 18.74 & -0.708 & 54.85
& 25.81 & 6.25 & 26.52 & 0.278
& 20.49 & 2.25 & 1.755 \\
MagicQuillV2
& 22.43 & \textbf{6.44} & 19.07 & -0.758 & 51.63
& 26.68 & \textbf{6.46} & 26.76 & 0.421
& 21.75 & 1.69 & 1.527 \\
\textbf{Ours}
& \textbf{22.44} & 6.39 & \textbf{19.88} & \textbf{-0.681} & \textbf{51.16}
& \textbf{26.76} & 6.40 & \textbf{26.91} & \textbf{0.493}
& $\infty$ & 0 & 0 \\
\bottomrule
\end{tabular*}
\end{table*}

\subsection{Quantitative Result}
Table~\ref{tab:all_settings} reports quantitative results for single-layer and multi-layer evaluation settings.
In the single-layer setting, LimeCross achieves the best overall edit-quality scores among the existing methods.
We note that preservation on strictly non-target content is guaranteed by design for LimeCross, as non-edited layers are kept unchanged by construction.
Under this definition, preservation can become trivial (e.g., PSNR$=\infty$, LPIPS$=0$, MSE$=0$).
We report these values for completeness and do not interpret them as evidence of superior edit quality.
In the multi-layer setting, LimeCross remains stable under sequential edits and yields stronger overall results, while baselines tend to degrade due to compounded errors from repeated flattening and ambiguity under occlusion.

\subsection{Ablation Study}
\label{subsec:ablation}

As shown in Fig.~\ref{fig:ablation}, we analyze the effect of the switching ratio $\rho \in [0,1]$.
Here, $\rho=0$ corresponds to always target-only sampling (Ours without context conditioning), while $\rho=1$ corresponds to always bi-stream conditioning (Ours without late-stage refinement).

We observe a clear trade-off between overall image quality and alpha-mask stability.
When $\rho$ is small, the model switches earlier to target-only refinement, limiting late-stage cross-layer interactions.
This yields better alpha stability and layer purity (lower mFID-$\alpha$), but leads to worse cross-layer coherence and thus worse overall image quality.
In contrast, when $\rho$ is large, bi-stream conditioning remains active longer, enabling stronger cross-layer interaction.
This yields better cross-layer coherence and overall image quality, but prolonged interaction during late refinement over-imposes contextual appearance onto the edited layer,
resulting in worse alpha stability (higher mFID-$\alpha$).
Overall, $\rho=0.5$ provides a balanced compromise.

\subsection{User Study}
Beyond automatic metrics, we conduct a user study to assess subjective human preferences in layered editing.
We compare LimeCross against MagicQuillV2~\cite{liu2025magicquillv2} under both single-layer and multi-layer settings.
We randomly sample 50 editing instances from LayerEditBench and evaluate them using a two-alternative forced-choice protocol.
30 participants are shown two outputs in randomized left--right order along with the target instruction.
To help participants focus on the intended edit without revealing method identity,
we display (i) the full recomposited image and (ii) the individual RGBA layer images corresponding to the edited result.
Participants answer four questions:
(1) which result looks more visually realistic and integrates the edit more naturally into the scene (image quality),
(2) which result better preserves the non-edited content (background preservation),
(3) which result better follows the target instruction (prompt adherence),
and (4) which result is preferred overall.

\begin{figure}[t]
\centering

\begin{minipage}[t]{0.43\linewidth}
  \vspace{0pt}
  \centering
  \includegraphics[width=\linewidth]{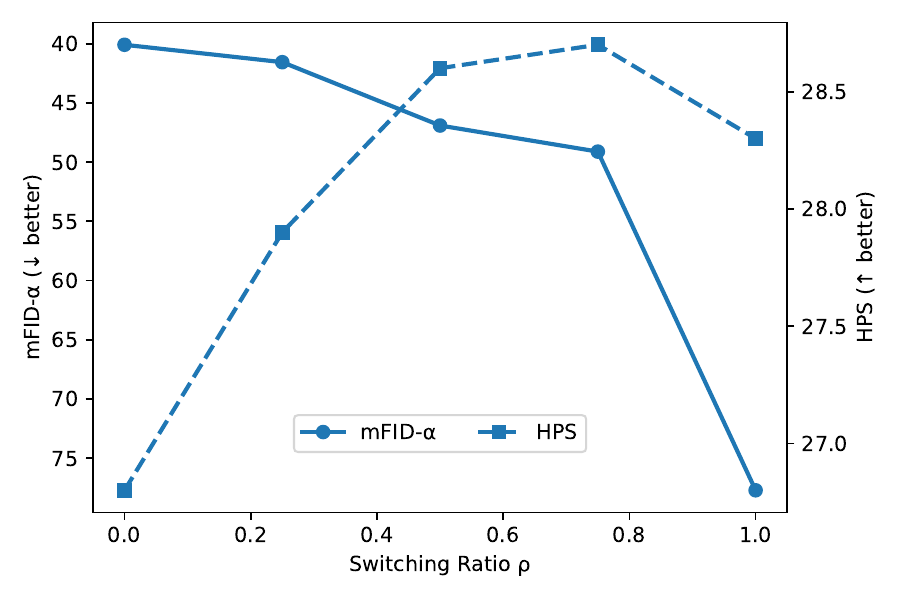}
  \caption{\textbf{Switching-ratio ablation.}
  Smaller $\rho$ yields better alpha stability but worse fidelity, while larger $\rho$ exhibits the opposite trend.}
  \label{fig:ablation}
\end{minipage}
\hfill
\begin{minipage}[t]{0.53\linewidth}
  \vspace{0pt}
  \centering
  \captionof{table}{User study. Preference rate (\%) of LimeCross over MagicQuillV2 with 95\% bootstrap confidence intervals. All results are significant under a two-sided binomial test ($p<0.001$).}
  \label{tab:userstudy}
  \small
  \renewcommand{\arraystretch}{1.15}

  \resizebox{\linewidth}{!}{%
  \begin{tabular}{lcc}
  \toprule
  Question & Pref.\%$\uparrow$ & 95\% CI \\
  \midrule
  Image quality & 67.3 & [64.0, 70.7] \\  
  Background preservation & 75.4 & [72.4, 78.5] \\ 
  Prompt adherence & 75.5 & [72.5, 78.6] \\ 
  Overall preference & 71.2 & [68.0, 74.4] \\ 
  \bottomrule
  \end{tabular}}
\end{minipage}

\end{figure}

We report the preference rate of LimeCross over MagicQuillV2 for each question, with 95\% confidence intervals computed by bootstrap resampling, and assess significance with a two-sided binomial test.
As summarized in Tab.~\ref{tab:userstudy}, LimeCross is preferred over MagicQuillV2 across the evaluated criteria.

\section{Conclusion}
\label{sec:conclusion}

We presented LimeCross, a training-free framework for context-conditioned layered image editing that preserves explicit RGBA structure during text-guided modification.
By enabling a selected layer to attend to contextual cues from remaining layers without flattening or re-decomposition, LimeCross maintains layer integrity and cross-layer consistency.
The proposed two-stage schedule further balances contextual alignment and alpha stability, supporting stable iterative multi-layer editing.
To facilitate systematic evaluation, we introduced LayerEditBench, a benchmark designed to assess both edit fidelity and layer purity.
Overall, our results highlight that treating layered editing as a structural problem, rather than editing flattened composites followed by decomposition, provides a foundation for controllable and reusable generative creation.

\bibliographystyle{splncs04}
\bibliography{main}

@String(CVPR  = {IEEE Conf. Comput. Vis. Pattern Recog.})

@String(ICCV  = {Int. Conf. Comput. Vis.})

@String(ECCV  = {Eur. Conf. Comput. Vis.})

@String(NeurIPS = {Adv. Neural Inform. Process. Syst.})

@String(ICML  = {Int. Conf. Mach. Learn.})

@String(ICASSP=	{ICASSP})

@String(CVPR  = {CVPR})

@String(ICCV  = {ICCV})

@String(ECCV  = {ECCV})

@String(WACV  = {WACV})

@String(NeurIPS = {NeurIPS})

@String(ICML  = {ICML})

@inproceedings{esser2024scaling,
  title={Scaling rectified flow transformers for high-resolution image synthesis},
  author={Esser, Patrick and Kulal, Sumith and Blattmann, Andreas and Entezari, Rahim and M{\"u}ller, Jonas and Saini, Harry and Levi, Yam and Lorenz, Dominik and Sauer, Axel and Boesel, Frederic and others},
  booktitle=ICML,
  year=2024
}

@inproceedings{quattrini2024alfie,
  title={Alfie: Democratising RGBA Image Generation with No},
  author={Quattrini, Fabio and Pippi, Vittorio and Cascianelli, Silvia and Cucchiara, Rita},
  booktitle=ECCV,
  pages={38--55},
  year=2024,
  organization={Springer}
}

@article{zhang2024transparent,
  title={Transparent image layer diffusion using latent transparency},
  author={Zhang, Lvmin and Agrawala, Maneesh},
  journal={arXiv preprint arXiv:2402.17113},
  year={2024}
}

@inproceedings{morita2025tkg,
  title={TKG-DM: Training-free Chroma Key Content Generation Diffusion Model},
  author={Morita, Ryugo and Frolov, Stanislav and Moser, Brian Bernhard and Shirakawa, Takahiro and Watanabe, Ko and Dengel, Andreas and Zhou, Jinjia},
  booktitle=CVPR,
  pages={13031--13040},
  year=2025
}

@article{nagai2025taue,
  title={TAUE: Training-free Noise Transplant and Cultivation Diffusion Model},
  author={Nagai, Daichi and Morita, Ryugo and Kitada, Shunsuke and Iyatomi, Hitoshi},
  journal={arXiv preprint arXiv:2511.02580},
  year={2025}
}

@article{huang2025dreamlayer,
  title={DreamLayer: Simultaneous Multi-Layer Generation via Diffusion Mode},
  author={Huang, Junjia and Yan, Pengxiang and Cai, Jinhang and Liu, Jiyang and Wang, Zhao and Wang, Yitong and Wu, Xinglong and Li, Guanbin},
  journal={arXiv preprint arXiv:2503.12838},
  year={2025}
}

@article{chen2025prismlayers,
  title={PrismLayers: Open Data for High-Quality Multi-Layer Transparent Image Generative Models},
  author={Chen, Junwen and Jiang, Heyang and Wang, Yanbin and Wu, Keming and Li, Ji and Zhang, Chao and Yanai, Keiji and Chen, Dong and Yuan, Yuhui},
  journal={arXiv preprint arXiv:2505.22523},
  year={2025}
}

@article{liu2025omnipsd,
  title={OmniPSD: Layered PSD Generation with Diffusion Transformer},
  author={Liu, Cheng and Song, Yiren and Wang, Haofan and Shou, Mike Zheng},
  journal={arXiv preprint arXiv:2512.09247},
  year={2025}
}

@article{chen2025inpainting,
  title={From Inpainting to Layer Decomposition: Repurposing Generative Inpainting Models for Image Layer Decomposition},
  author={Chen, Jingxi and Zhang, Yixiao and Qian, Xiaoye and Li, Zongxia and Fermuller, Cornelia and Chen, Caren and Aloimonos, Yiannis},
  journal={arXiv preprint arXiv:2511.20996},
  year={2025}
}

@inproceedings{ao2025open,
  title={Open-world amodal appearance completion},
  author={Ao, Jiayang and Jiang, Yanbei and Ke, Qiuhong and Ehinger, Krista A},
  booktitle=CVPR,
  pages={6490--6499},
  year={2025}
}

@inproceedings{dalva2024layerfusion,
  title={Layerfusion: Harmonized multi-layer text-to-image generation with generative priors},
  author={Dalva, Yusuf and Li, Yijun and Liu, Qing and Zhao, Nanxuan and Zhang, Jianming and Lin, Zhe and Yanardag, Pinar},
  booktitle={NeurIPS 2025 Workshop on Space in Vision, Language, and Embodied AI},
  year={2024}
}

@article{yin2025qwen,
  title={Qwen-Image-Layered: Towards Inherent Editability via Layer Decomposition},
  author={Yin, Shengming and Zhang, Zekai and Tang, Zecheng and Gao, Kaiyuan and Xu, Xiao and Yan, Kun and Li, Jiahao and Chen, Yilei and Chen, Yuxiang and Shum, Heung-Yeung and others},
  journal={arXiv preprint arXiv:2512.15603},
  year={2025}
}

@article{fontanella2024generating,
  title={Generating compositional scenes via Text-to-image RGBA Instance Generation},
  author={Fontanella, Alessandro and Tudosiu, Petru-Daniel and Yang, Yongxin and Zhang, Shifeng and Parisot, Sarah},
  journal=NeurIPS,
  volume={37},
  pages={43864--43893},
  year={2024}
}

@inproceedings{huang2024layerdiff,
  title={Layerdiff: Exploring text-guided multi-layered composable image synthesis via layer-collaborative diffusion model},
  author={Huang, Runhui and Cai, Kaixin and Han, Jianhua and Liang, Xiaodan and Pei, Renjing and Lu, Guansong and Xu, Songcen and Zhang, Wei and Xu, Hang},
  booktitle=ECCV,
  pages={144--160},
  year={2024},
  organization={Springer}
}

@inproceedings{niedecomposition,
  title={Decomposition of Graphic Design with Unified Multimodal Model},
  author={Nie, Hui and Zhang, Zhao and Cheng, Yutao and Yang, Maoke and Shi, Gonglei and Xie, Qingsong and Shao, Jie and Wu, Xinglong},
  booktitle=ICML,
  year=2025,
}

@article{kang2025layeringdiff,
  title={LayeringDiff: Layered Image Synthesis via Generation, then Disassembly with Generative Knowledge},
  author={Kang, Kyoungkook and Sim, Gyujin and Kim, Geonung and Kim, Donguk and Nam, Seungho and Cho, Sunghyun},
  journal={arXiv preprint arXiv:2501.01197},
  year={2025}
}

@article{huang2025psdiffusion,
  title={Psdiffusion: Harmonized multi-layer image generation via layout and appearance alignment},
  author={Huang, Dingbang and Li, Wenbo and Zhao, Yifei and Pan, Xinyu and Wang, Chun and Zeng, Yanhong and Dai, Bo},
  journal={arXiv preprint arXiv:2505.11468},
  year={2025}
}

@inproceedings{yang2025generative,
  title={Generative Image Layer Decomposition with Visual Effects},
  author={Yang, Jinrui and Liu, Qing and Li, Yijun and Kim, Soo Ye and Pakhomov, Daniil and Ren, Mengwei and Zhang, Jianming and Lin, Zhe and Xie, Cihang and Zhou, Yuyin},
  booktitle=CVPR,
  pages={7643--7653},
  year={2025}
}

@inproceedings{zou2025zero,
  title={Zero-Shot Subject-Centric Generation for Creative Application Using Entropy Fusion},
  author={Zou, Kaifeng and Feng, Xiaoyi and Huang, Tao and Huang, Zizhou and Zhang, Haihang and Zou, Yuntao and Li, Dagang},
  booktitle=ICCV,
  pages={6136--6145},
  year={2025}
}

@inproceedings{pu2025art,
  title={Art: Anonymous region transformer for variable multi-layer transparent image generation},
  author={Pu, Yifan and Zhao, Yiming and Tang, Zhicong and Yin, Ruihong and Ye, Haoxing and Yuan, Yuhui and Chen, Dong and Bao, Jianmin and Zhang, Sirui and Wang, Yanbin and others},
  booktitle=CVPR,
  pages={7952--7962},
  year={2025}
}

@inproceedings{tudosiu2024mulan,
  title={Mulan: A multi layer annotated dataset for controllable text-to-image generation},
  author={Tudosiu, Petru-Daniel and Yang, Yongxin and Zhang, Shifeng and Chen, Fei and McDonagh, Steven and Lampouras, Gerasimos and Iacobacci, Ignacio and Parisot, Sarah},
  booktitle=CVPR,
  pages={22413--22422},
  year={2024}
}

@article{zhang2023text2layer,
  title={Text2layer: Layered image generation using latent diffusion model},
  author={Zhang, Xinyang and Zhao, Wentian and Lu, Xin and Chien, Jeff},
  journal={arXiv preprint arXiv:2307.09781},
  year={2023}
}

@article{liu2025controllable,
  title={Controllable Layer Decomposition for Reversible Multi-Layer Image Generation},
  author={Liu, Zihao and Xu, Zunnan and Shu, Shi and Zhou, Jun and Zhang, Ruicheng and Tang, Zhenchao and Li, Xiu},
  journal={arXiv preprint arXiv:2511.16249},
  year={2025}
}

@inproceedings{dai2025trans,
  title={Trans-adapter: A plug-and-play framework for transparent image inpainting},
  author={Dai, Yuekun and Li, Haitian and Zhou, Shangchen and Loy, Chen Change},
  booktitle=ICCV,
  pages={15015--15024},
  year={2025}
}

@article{liu2025magicquillv2,
  title={MagicQuillV2: Precise and Interactive Image Editing with Layered Visual Cues},
  author={Liu, Zichen and Yu, Yue and Ouyang, Hao and Wang, Qiuyu and Ma, Shuailei and Cheng, Ka Leong and Wang, Wen and Bai, Qingyan and Zhang, Yuxuan and Zeng, Yanhong and others},
  journal={arXiv preprint arXiv:2512.03046},
  year={2025}
}

@article{song2025layertracer,
  title={Layertracer: Cognitive-aligned layered svg synthesis via diffusion transformer},
  author={Song, Yiren and Chen, Danze and Shou, Mike Zheng},
  journal={arXiv preprint arXiv:2502.01105},
  year={2025}
}

@article{rout2024semantic,
  title={Semantic image inversion and editing using rectified stochastic differential equations},
  author={Rout, Litu and Chen, Yujia and Ruiz, Nataniel and Caramanis, Constantine and Shakkottai, Sanjay and Chu, Wen-Sheng},
  journal={arXiv preprint arXiv:2410.10792},
  year={2024}
}

@article{wang2024taming,
  title={Taming rectified flow for inversion and editing},
  author={Wang, Jiangshan and Pu, Junfu and Qi, Zhongang and Guo, Jiayi and Ma, Yue and Huang, Nisha and Chen, Yuxin and Li, Xiu and Shan, Ying},
  journal={arXiv preprint arXiv:2411.04746},
  year={2024}
}

@inproceedings{xu2025unveil,
  title={Unveil inversion and invariance in flow transformer for versatile image editing},
  author={Xu, Pengcheng and Jiang, Boyuan and Hu, Xiaobin and Luo, Donghao and He, Qingdong and Zhang, Jiangning and Wang, Chengjie and Wu, Yunsheng and Ling, Charles and Wang, Boyu},
  booktitle=CVPR,
  pages={28479--28489},
  year={2025}
}

@article{yan2025eedit,
  title={Eedit: Rethinking the spatial and temporal redundancy for efficient image editing},
  author={Yan, Zexuan and Ma, Yue and Zou, Chang and Chen, Wenteng and Chen, Qifeng and Zhang, Linfeng},
  journal={arXiv preprint arXiv:2503.10270},
  year={2025}
}

@article{xie2025dnaedit,
  title={DNAEdit: Direct Noise Alignment for Text-Guided Rectified Flow Editing},
  author={Xie, Chenxi and Li, Minghan and Li, Shuai and Wu, Yuhui and Yi, Qiaosi and Zhang, Lei},
  journal={arXiv preprint arXiv:2506.01430},
  year={2025}
}

@article{ma2025adams,
  title={Adams Bashforth Moulton Solver for Inversion and Editing in Rectified Flow},
  author={Ma, Yongjia and Di, Donglin and Liu, Xuan and Chen, Xiaokai and Fan, Lei and Chen, Wei and Su, Tonghua},
  journal={arXiv preprint arXiv:2503.16522},
  year={2025}
}

@article{ronai2025flowopt,
  title={FlowOpt: Fast Optimization Through Whole Flow Processes for Training-Free Editing},
  author={Ronai, Or and Kulikov, Vladimir and Michaeli, Tomer},
  journal={arXiv preprint arXiv:2510.22010},
  year={2025}
}

@article{xu2023inversion,
  title={Inversion-free image editing with natural language},
  author={Xu, Sihan and Huang, Yidong and Pan, Jiayi and Ma, Ziqiao and Chai, Joyce},
  journal={arXiv preprint arXiv:2312.04965},
  year={2023}
}

@inproceedings{kulikov2025flowedit,
  title={Flowedit: Inversion-free text-based editing using pre-trained flow models},
  author={Kulikov, Vladimir and Kleiner, Matan and Huberman-Spiegelglas, Inbar and Michaeli, Tomer},
  booktitle=ICCV,
  pages={19721--19730},
  year={2025}
}

@article{kim2025flowalign,
  title={Flowalign: Trajectory-regularized, inversion-free flow-based image editing},
  author={Kim, Jeongsol and Hong, Yeobin and Park, Jonghyun and Ye, Jong Chul},
  journal={arXiv preprint arXiv:2505.23145},
  year={2025}
}

@article{yang2025fia,
  title={FIA-Edit: Frequency-Interactive Attention for Efficient and High-Fidelity Inversion-Free Text-Guided Image Editing},
  author={Yang, Kaixiang and Shen, Boyang and Li, Xin and Dai, Yuchen and Luo, Yuxuan and Ma, Yueran and Fang, Wei and Li, Qiang and Wang, Zhiwei},
  journal={arXiv preprint arXiv:2511.12151},
  year={2025}
}

@article{zhu2025kv,
  title={Kv-edit: Training-free image editing for precise background preservation},
  author={Zhu, Tianrui and Zhang, Shiyi and Shao, Jiawei and Tang, Yansong},
  journal={arXiv preprint arXiv:2502.17363},
  year={2025}
}

@article{ouyang2025lore,
  title={LORE: Latent Optimization for Precise Semantic Control in Rectified Flow-based Image Editing},
  author={Ouyang, Liangyang and Mao, Jiafeng},
  journal={arXiv preprint arXiv:2508.03144},
  year={2025}
}

@inproceedings{suzuki2025layerd,
  title={Layerd: Decomposing raster graphic designs into layers},
  author={Suzuki, Tomoyuki and Liu, Kang-Jun and Inoue, Naoto and Yamaguchi, Kota},
  booktitle=ICCV,
  pages={17783--17792},
  year={2025}
}

@article{wang2025alphavae,
  title={Alphavae: Unified end-to-end RGBA image reconstruction and generation with alpha-aware representation learning},
  author={Wang, Zile and Yu, Hao and Zhan, Jiabo and Yuan, Chun},
  journal={arXiv preprint arXiv:2507.09308},
  year={2025}
}

@misc{flux2025,
  title        = {FLUX},
  author       = {Black Forest Labs},
  howpublished = {\url{https://github.com/black-forest-labs/flux}},
  year         = {2025}
}

@article{dong2025video,
  title={Video Generation with Stable Transparency via Shiftable RGB-A Distribution Learner},
  author={Dong, Haotian and Wang, Wenjing and Li, Chen and Lyu, Jing and Lin, Di},
  journal={arXiv preprint arXiv:2509.24979},
  year={2025}
}

@inproceedings{ji2025layerflow,
  title={Layerflow: A unified model for layer-aware video generation},
  author={Ji, Sihui and Luo, Hao and Chen, Xi and Tu, Yuanpeng and Wang, Yiyang and Zhao, Hengshuang},
  booktitle={Proceedings of the Special Interest Group on Computer Graphics and Interactive Techniques Conference Conference Papers},
  pages={1--10},
  year={2025}
}

@inproceedings{wang2025transpixeler,
  title={TransPixeler: Advancing Text-to-Video Generation with Transparency},
  author={Wang, Luozhou and Li, Yijun and Chen, Zhifei and Wang, Jui-Hsien and Zhang, Zhifei and Zhang, He and Lin, Zhe and Chen, Ying-Cong},
  booktitle=CVPR,
  pages={18229--18239},
  year={2025}
}

@article{chen2025transanimate,
  title={Transanimate: Taming layer diffusion to generate rgba video},
  author={Chen, Xuewei and Chen, Zhimin and Song, Yiren},
  journal={arXiv preprint arXiv:2503.17934},
  year={2025}
}

@article{cen2025layert2v,
  title={LayerT2V: Interactive Multi-Object Trajectory Layering for Video Generation},
  author={Cen, Kangrui and Zhao, Baixuan and Xin, Yi and Luo, Siqi and Zhai, Guangtao and Liu, Xiaohong},
  journal={arXiv preprint arXiv:2508.04228},
  year={2025}
}

@inproceedings{bai2025layer,
  title={Layer-Animate for Transparent Video Generation},
  author={Bai, Jingqi and Zhou, Jingkai and Wang, Benzhi and Chen, Weihua and Yang, Yang and Lei, Zhen and Wang, Fan},
  booktitle={ICASSP 2025-2025 IEEE International Conference on Acoustics, Speech and Signal Processing (ICASSP)},
  pages={1--5},
  year={2025},
  organization={IEEE}
}

@inproceedings{lee2025generative,
  title={Generative omnimatte: Learning to decompose video into layers},
  author={Lee, Yao-Chih and Lu, Erika and Rumbley, Sarah and Geyer, Michal and Huang, Jia-Bin and Dekel, Tali and Cole, Forrester},
  booktitle=CVPR,
  pages={12522--12532},
  year={2025}
}

@article{achiam2023gpt,
  title={Gpt-4 technical report},
  author={Achiam, Josh and Adler, Steven and Agarwal, Sandhini and Ahmad, Lama and Akkaya, Ilge and Aleman, Florencia Leoni and Almeida, Diogo and Altenschmidt, Janko and Altman, Sam and Anadkat, Shyamal and others},
  journal={arXiv preprint arXiv:2303.08774},
  year={2023}
}

@article{hertz2022prompt,
  title={Prompt-to-prompt image editing with cross attention control},
  author={Hertz, Amir and Mokady, Ron and Tenenbaum, Jay and Aberman, Kfir and Pritch, Yael and Cohen-Or, Daniel},
  journal={arXiv preprint arXiv:2208.01626},
  year={2022}
}

@article{ju2023direct,
  title={Direct inversion: Boosting diffusion-based editing with 3 lines of code},
  author={Ju, Xuan and Zeng, Ailing and Bian, Yuxuan and Liu, Shaoteng and Xu, Qiang},
  journal={arXiv preprint arXiv:2310.01506},
  year={2023}
}

@inproceedings{mokady2023null,
  title={Null-text inversion for editing real images using guided diffusion models},
  author={Mokady, Ron and Hertz, Amir and Aberman, Kfir and Pritch, Yael and Cohen-Or, Daniel},
  booktitle=CVPR,
  pages={6038--6047},
  year={2023}
}

@inproceedings{morita2023interactive,
  title={Interactive image manipulation with complex text instructions},
  author={Morita, Ryugo and Zhang, Zhiqiang and Ho, Man M and Zhou, Jinjia},
  booktitle=WACV,
  pages={1053--1062},
  year={2023}
}

@article{wu2023human,
  title={Human preference score v2: A solid benchmark for evaluating human preferences of text-to-image synthesis},
  author={Wu, Xiaoshi and Hao, Yiming and Sun, Keqiang and Chen, Yixiong and Zhu, Feng and Zhao, Rui and Li, Hongsheng},
  journal={arXiv preprint arXiv:2306.09341},
  year={2023}
}

@article{schuhmann2022laion,
  title={Laion-5b: An open large-scale dataset for training next generation image-text models},
  author={Schuhmann, Christoph and Beaumont, Romain and Vencu, Richard and Gordon, Cade and Wightman, Ross and Cherti, Mehdi and Coombes, Theo and Katta, Aarush and Mullis, Clayton and Wortsman, Mitchell and others},
  journal=NeurIPS,
  volume={35},
  pages={25278--25294},
  year={2022}
}

@inproceedings{hessel2021clipscore,
  title={Clipscore: A reference-free evaluation metric for image captioning},
  author={Hessel, Jack and Holtzman, Ari and Forbes, Maxwell and Le Bras, Ronan and Choi, Yejin},
  booktitle={Proceedings of the 2021 conference on empirical methods in natural language processing},
  pages={7514--7528},
  year={2021}
}

@article{xu2023imagereward,
  title={Imagereward: Learning and evaluating human preferences for text-to-image generation},
  author={Xu, Jiazheng and Liu, Xiao and Wu, Yuchen and Tong, Yuxuan and Li, Qinkai and Ding, Ming and Tang, Jie and Dong, Yuxiao},
  journal=NeurIPS,
  volume={36},
  pages={15903--15935},
  year={2023}
}

@article{labs2025flux,
  title={FLUX. 1 Kontext: Flow Matching for In-Context Image Generation and Editing in Latent Space},
  author={Labs, Black Forest and Batifol, Stephen and Blattmann, Andreas and Boesel, Frederic and Consul, Saksham and Diagne, Cyril and Dockhorn, Tim and English, Jack and English, Zion and Esser, Patrick and others},
  journal={arXiv preprint arXiv:2506.15742},
  year={2025}
}

@inproceedings{brooks2023instructpix2pix,
  title={Instructpix2pix: Learning to follow image editing instructions},
  author={Brooks, Tim and Holynski, Aleksander and Efros, Alexei A},
  booktitle=CVPR,
  pages={18392--18402},
  year={2023}
}

@misc{qwenlayeredcontrol2025,
  title={Qwen-Image-Layered-Control},
  author={DiffSynth-Studio},
  year={2025},
  howpublished={\url{https://huggingface.co/DiffSynth-Studio/Qwen-Image-Layered-Control}}
}

@article{morita2026lgtm,
  title={LGTM: Training-Free Light-Guided Text-to-Image Diffusion Model via Initial Noise Manipulation},
  author={Morita, Ryugo and Frolov, Stanislav and Moser, Brian Bernhard and Watanabe, Ko and Takahashi, Riku and Dengel, Andreas},
  journal={arXiv preprint arXiv:2603.24086},
  year={2026}
}

\end{document}